\documentclass[conference]{IEEEtran}
\IEEEoverridecommandlockouts
\usepackage[T1]{fontenc}

\usepackage{cite}
\usepackage{times}
\usepackage{epsfig}
\usepackage{graphicx}
\usepackage{amsmath,amssymb,amsfonts}
\usepackage{booktabs}
\usepackage{multirow}
\usepackage{multicol}
\usepackage{array}
\usepackage{xcolor}
\usepackage{colortbl}
\usepackage{subfigure}
\usepackage{algorithm}
\usepackage{algorithmic}
\usepackage{enumitem}
\usepackage{balance}
\usepackage{threeparttable}
\usepackage{textcomp}
\usepackage{enumitem}
\usepackage[margin=1in]{geometry}
\usepackage[breaklinks=true,colorlinks,bookmarks=false]{hyperref}
\usepackage{tabularx}

\newcommand{\method}{ViThinker}

\newcolumntype{L}[1]{>{\raggedright\arraybackslash}p{#1}}
\newcolumntype{C}[1]{>{\centering\arraybackslash}p{#1}}
\newcolumntype{R}[1]{>{\raggedleft\arraybackslash}p{#1}}

\usepackage{wrapfig}

\usepackage[most]{tcolorbox}
\tcbuselibrary{skins, breakable}

\definecolor{querycolor}{RGB}{255,182,193}
\definecolor{thinkcolor}{RGB}{230,242,255}
\definecolor{answercolor}{RGB}{230,255,230}
\definecolor{headerseq}{RGB}{100,100,100}
\definecolor{headervit}{RGB}{46,125,50}

\newcolumntype{C}{>{\centering\arraybackslash}X}

\tcbset{
    mystagebox/.style={
    enhanced,boxrule=0.8pt,coltitle=white,fonttitle=\bfseries,sharp corners=south, 
    }}

\def\BibTeX{{\rm B\kern-.05em{\sc i\kern-.025em b}\kern-.08em
    T\kern-.1667em\lower.7ex\hbox{E}\kern-.125emX}}
    
\begin{document}
\title{\method{}: Active Vision-Language Reasoning via Dynamic Perceptual Querying}

\author{
\IEEEauthorblockN{
Weihang You$^{*,1}$,
Qingchan Zhu$^{*,1}$,
David Liu$^{*,2}$,
Yi Pan$^{1}$,
Geng Yuan$^{1}$,
Hanqi Jiang$^{\dagger, 1}$
}
\IEEEauthorblockA{
$^{1}$School of Computing, University of Georgia, Athens, GA, USA\\
$^{2}$School of Engineering and Applied Science, Princeton University, Princeton, NJ, USA\\
}
\thanks{$^{*}$Co-first authors.} 
\thanks{$^{\dagger}$Corresponding Author: Hanqi Jiang (hanqi.jiang@uga.edu).}}

\maketitle


\begin{abstract}
Chain-of-Thought (CoT) reasoning excels in language models but struggles in vision-language models due to premature visual-to-text conversion that discards continuous information such as geometry and spatial layout. While recent methods enhance CoT through static enumeration or attention-based selection, they remain passive, i.e., processing pre-computed inputs rather than actively seeking task-relevant details. Inspired by human active perception, we introduce ViThinker, a framework that enables vision-language models to autonomously generate decision (query) tokens triggering the synthesis of expert-aligned visual features on demand. ViThinker internalizes vision-expert capabilities during training, performing generative mental simulation during inference without external tool calls. Through a two-stage curriculum: first distilling frozen experts into model parameters, then learning task-driven querying via sparsity penalties, i.e., ViThinker discovers minimal sufficient perception for each reasoning step. Evaluations across vision-centric benchmarks demonstrate consistent improvements, validating that active query generation outperforms passive approaches in both perceptual grounding and reasoning accuracy.
\end{abstract}


\section{Introduction}

Vision-Language Models (VLMs)~\cite{bai2025qwen25vltechnicalreport, liu2023llava} have achieved remarkable progress in multimodal understanding. A key enabler of this success is Chain-of-Thought (CoT) reasoning~\cite{wei2023chainofthought}, which decomposes complex problems into intermediate steps, significantly improving model performance. However, while CoT excels in text-only language models, it faces a fundamental limitation when applied to VLMs: \textit{premature visual-to-text conversion}. Purely textual CoT forces models to verbalize visual observations early in the reasoning process, discarding continuous visual information such as precise geometry, spatial layout, object boundaries, and fine-grained structure. This text-biased approach struggles with tasks requiring grounded perceptual details, where critical visual signals are lost in translation.

Recent efforts have sought to address this by incorporating richer visual representations into reasoning chains. Methods like Aurora~\cite{bigverdi2024perceptiontokensenhancevisual} generates perception tokens, ICoT~\cite{gao2025interleaved} selects features via attention, and CoVT~\cite{qin2025covt} enumerates dense visual features. Yet these approaches remain fundamentally \textit{passive}: they process static, pre-computed visual inputs without actively seeking task-relevant details. This leads to either noisy, inefficient reasoning from over-enumeration or insufficient perceptual precision from passive selection.

In contrast, human perception is inherently \textit{active}~\cite{clark2013whatever}. When solving visual problems, we strategically decide \textit{how to see} and \textit{when to look}, mentally simulating specific perceptual cues (e.g., contour tracing, depth estimation) only when reasoning demands them. This metacognitive flexibility enables grounded problem-solving. Inspired by this, we introduce \textbf{ViThinker}, a framework for active vision-language reasoning via dynamic perceptual querying. ViThinker enables models to autonomously generate decision (query) tokens (e.g., <query\_depth>, <query\_seg>) that trigger the synthesis of task-relevant visual features on demand. Unlike tool-use agents that invoke external tool calls during inference~\cite{schick2023toolformer}, ViThinker internalizes vision experts during training by distilling frozen experts into model parameters. During inference, the model performs generative mental simulation, synthesizing expert-aligned features from parametric memory to establish a "Think-Query-Simulate-Think" loop. 

To train this capability, we propose a two-stage curriculum: the model first learns \textit{how to see} (distilling vision experts' knowledge), then learns \textit{when to look} (discovering task-relevant queries). A sparsity penalty is added to the loss function to enforce a cognitive budget, guiding the model toward minimal sufficient perception.

Through this design, ViThinker bridges the gap between passive feature processing and active perceptual reasoning, enabling VLMs to exhibit human-like metacognitive flexibility in visual problem-solving.\\ 
Our contributions are as follows:
\begin{itemize}[leftmargin=*, nosep]
    \item We propose \textbf{ViThinker}, a paradigm shifting from passive feature processing to active perceptual simulation via task-driven query generation.
    \item We design a \textbf{Decoupled Query Mechanism} separating strategic intent (Decision Tokens) from generative execution (Observation Tokens), and a \textbf{Strategic Policy Learning curriculum} that teaches task-driven perceptual selection via sparsity penalties.
     \item We demonstrate \textbf{consistent improvements} in both perceptual grounding and reasoning accuracy across vision-centric benchmarks, validating that task-driven query generation outperforms passive approaches.
\end{itemize}

\begin{figure*}[t]
    \centering
    \includegraphics[width=\textwidth]{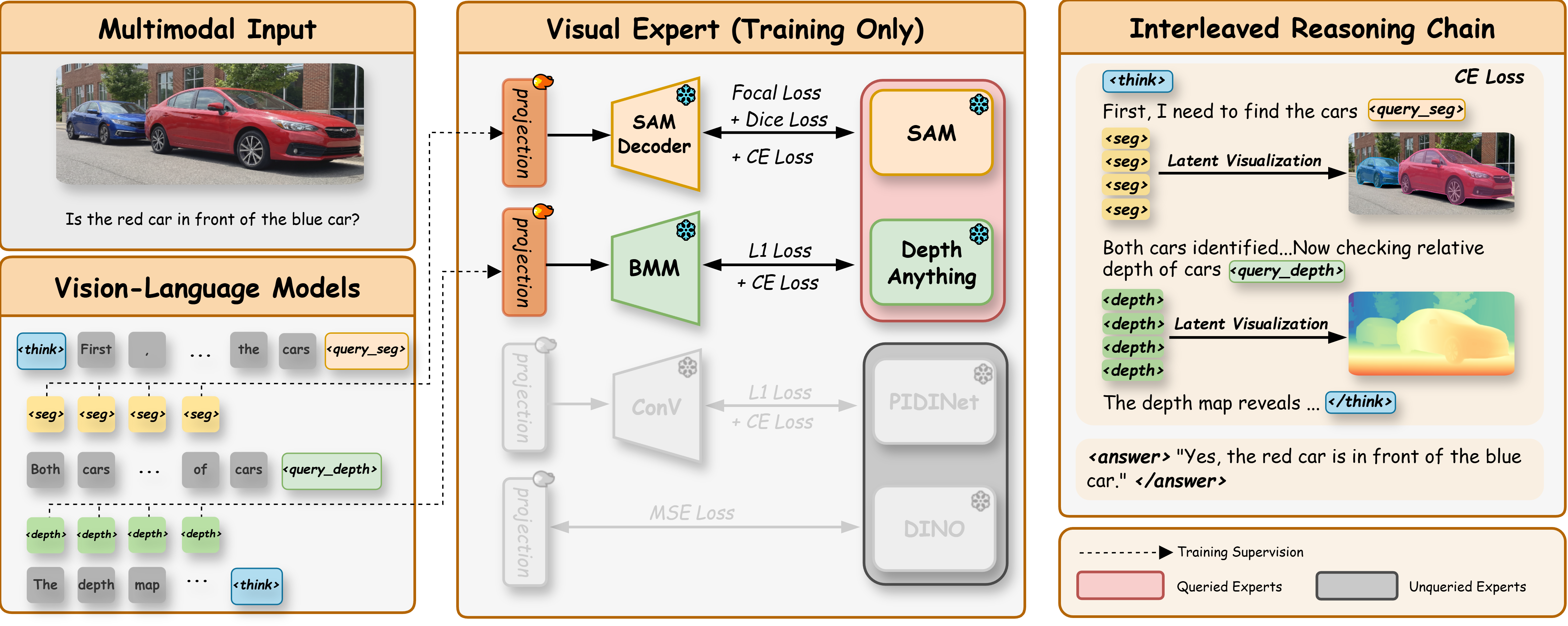}
    \caption{\textbf{Overview of ViThinker Framework.} ViThinker enables interleaved vision-language reasoning through a "Think-Query-Simulate-Think" loop. \textbf{Left}: Given a Visual-QA input, the VLM processes image and text into tokens. \textbf{Middle}: During training, we distill frozen vision experts (SAM, DepthAnything, PIDINet, DINOv2) into the VLM via multi-task supervision. \textbf{Right}: At inference, ViThinker performs internalized simulation. Crucially, no external vision models are invoked; instead, the generated query tokens (e.g., <query\_seg>) directly trigger the synthesis of expert-aligned features from the model's parameters.}
\label{fig:framework}
\end{figure*}

\section{Method}
In this section, we present ViThinker, a framework designed for active and interleaved vision-language reasoning (Fig~\ref{fig:framework}). We first introduce the Active Generative Perception Module which enables the model to autonomously simulate visual information (Sec.~\ref{sec:visual-query}). We then detail the Policy Learning Curriculum that trains the model to balance reasoning accuracy with token efficiency (Sec.~\ref{sec:training-stages}). Finally, we formalize the training objectives that drive this emergent behavior (Sec.~\ref{sec:training-objectives}).

\subsection{Active Generative Perception Module}
\label{sec:visual-query}
To transform standard VLMs from passive observers into active inquirers, we augment the vocabulary with a set of dedicated \textit{Decision Tokens}:
\begin{equation}
\label{eq:tokens}
\begin{aligned}
\mathcal{V}_{trig} = \{ & \text{<query\_seg>}, \text{<query\_depth>}, \\
                        & \text{<query\_edge>}, \text{<query\_patch>} \}
\end{aligned}
\end{equation}

These tokens serve as explicit cognitive actions. They allow the model to pause textual generation and autonomously initiate a perceptual simulation process to resolve ambiguity.

\subsubsection{Expert Internalization via Alignment} 
A core innovation of ViThinker is the internalization of expert capabilities. We decouple the "decision to look" from the "act of seeing." When the model generates a \textit{Decision Token} $\in \mathcal{V}_{trig}$, it triggers a specialized encoding phase. 

The subsequent four positions are reserved as \textit{Observation Tokens} $\mathbf{vis}_m = \{vis_m^{(1)}, vis_m^{(2)}, vis_m^{(3)}, vis_m^{(4)}\}$ and type $m \in \{\text{seg}, \text{depth}, \text{edge}, \text{patch}\}$. We align the hidden states $\mathbf{h}_{vis} = h_{t+1:t+4}$ corresponding to these observation tokens with the feature maps of frozen experts $\Phi_m(I)$ through a projection head $\text{Proj}_m$:
\begin{equation}
\label{eq:align}
\mathcal{L}_{align}^m = \mathcal{D}\left( \text{Proj}_m(\mathbf{h}_{vis}), \Phi_m(I) \right)
\end{equation}
where $\mathcal{D}$ denotes the expert-specific distance metric. Each $\text{Proj}_m$ follows a similar architecture but has different input/output dimensions: a linear layer projects the hidden states into the required expert space, and a learnable query cross-attends to these projected features (serving as both keys and values) to produce the final aligned embeddings.

We utilize four complementary experts: SAM~\cite{kirillov2023segment} for segmentation (seg), DepthAnything~\cite{yang2024depthv2} for geometry (depth), PIDINet~\cite{su2021pixeldifferencenetworksefficient} for structural edges (edge), and DINOv2~\cite{oquab2024dinov2learningrobustvisual} for patch level semantic correspondence (patch). This alignment effectively distills the experts' explicit knowledge into the VLM's parametric memory.

\subsubsection{Inference---Perceptual Simulation} 
Unlike tool-use agents that invoke external APIs, ViThinker performs \textit{internalized reasoning} through generative mental simulation. During inference, query tokens trigger the synthesis of expert-aligned visual features directly from the model's memory, which utilizes the alignment learned during training to reconstruct perceptual details. This process mirrors human perception: the model actively "recalls" specific perceptual cues (e.g., depth geometry, object boundaries) from raw visual features only when reasoning demands them.

\subsection{Policy Learning Curriculum}
\label{sec:training-stages}
Training a model to actively simulate perception requires more than simple supervision. We propose a two-stage curriculum designed first to internalize expert capabilities (\textit{how to see}), then learn strategic, task-driven querying (\textit{when to look}).

\subsubsection{Stage 1: Perceptual Skill Acquisition}
\begin{figure}[t]
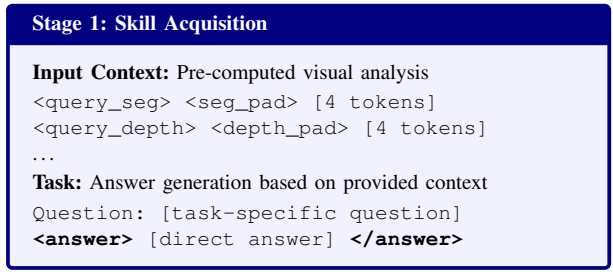

\centering
\footnotesize
\begin{tcolorbox}[colback=blue!5, colframe=blue!50!black, title={\footnotesize\textbf{Stage 1: Skill Acquisition}}, fonttitle=\bfseries, arc=1pt, left=2mm, right=2mm, top=1.5mm, bottom=1.5mm]
\textbf{Input Context:} Pre-computed visual analysis\\[2pt]
\texttt{<query\_seg> <seg\_pad> [4 tokens]} \\[1pt]
\texttt{<query\_depth> <depth\_pad> [4 tokens]}\\[1pt]
\dots \\[2pt]
\textbf{Task:} Answer generation based on provided context\\[2pt]
\texttt{Question: [task-specific question]}\\[1pt]
\texttt{\textbf{<answer>} [direct answer] \textbf{</answer>}}
\end{tcolorbox}
\caption{Stage 1 focuses on skill acquisition. The model learns to encode and synthesize visual features by observing expert outputs provided in the context.}
\label{fig:stage1_format}
\vspace{-2mm}
\end{figure}

The first stage establishes a foundation by internalizing expert knowledge about vision into model parameters. We construct a dataset where expert outputs are prepended to the input context. As shown in Fig.~\ref{fig:stage1_format}, each <query\_xxx> sequence is aligned with its corresponding expert features via supervision losses (Eq.~\ref{eq:align}). This stage teaches the model the semantic meaning of each Decision Token and how to synthesize high-fidelity visual features, effectively distilling expert capabilities into parametric memory.

\subsubsection{Stage 2: Strategic Policy Optimization}
\begin{figure}[t]
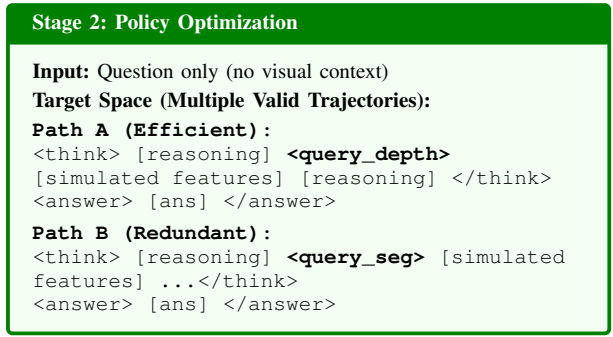

\centering
\footnotesize
\begin{tcolorbox}[colback=green!5, colframe=green!50!black, title={\footnotesize\textbf{Stage 2: Policy Optimization}}, fonttitle=\bfseries, arc=1pt, left=2mm, right=2mm, top=1.5mm, bottom=1.5mm]
\textbf{Input:} Question only (no visual context)\\[2pt]
\textbf{Target Space (Multiple Valid Trajectories):}\\[2pt]
\texttt{\textbf{Path A (Efficient)}:}\\
\texttt{<think> [reasoning] \textbf{<query\_depth>} [simulated features] [reasoning] </think>}\\
\texttt{<answer> [ans] </answer>}\\[3pt]
\texttt{\textbf{Path B (Redundant)}:}\\
\texttt{<think> [reasoning] \textbf{<query\_seg>} [simulated features] \dots </think>}\\
\texttt{<answer> [ans] </answer>}
\end{tcolorbox}
\caption{Stage 2 optimizes the decision policy. By presenting multiple valid reasoning paths with different perceptual coverage, the sparsity penalty guides the model toward task-appropriate feature selection.}
\label{fig:stage2_format}
\vspace{-4mm}
\end{figure}

The second stage shifts from passive feature processing to \textit{active, task-driven querying}. The model must learn \textit{which} perceptual simulations are necessary for each reasoning step to avoid both insufficient grounding from under-querying and conflicting perceptual signals from indiscriminate enumeration. To achieve this, we construct a training landscape with multiple valid reasoning paths for each problem (Fig.~\ref{fig:stage2_format}). Chain variants, ranging from minimal (single expert) to comprehensive (all experts), are generated by Gemini Flash~\cite{geminiteam2025} and are programmatically validated using task-specific constraints on decision tokens. The data distribution includes 20\% full coverage, 60\% task-specific subsets, and 20\% minimal queries. This diversity, combined with sparsity penalties (Sec.~\ref{sec:sparsity}), guides the model to actively select task-appropriate experts rather than passively enumerate.

\subsubsection{Sparsity-Driven Decision Making} 
\label{sec:sparsity}
To enforce a \textit{cognitive budget} and guide the model toward minimal sufficient perception, we apply a sparsity penalty to decision tokens:
\begin{equation}
\mathcal{L}_{p} = \sum_{t \in \mathcal{T}_{q}} \omega(Q_t), \quad \omega(Q_t) = N
\end{equation}
where $\mathcal{T}_{q}$ are decision token indices and $N$ is the number of Observation Tokens per decision token. This design is crucial for learning \textit{when to look}. By penalizing the \textit{decision} rather than the \textit{representation}, we decouple strategic selection from perceptual quality: observation tokens remain free to learn high-fidelity features via alignment loss (Eq.~\ref{eq:align}), while decision tokens absorb sparsity pressure. This guides the model to actively select task-relevant experts rather than passively enumerating them, thereby discovering the minimal sufficient perceptual simulation for each task.

\subsection{Training Objectives}
\label{sec:training-objectives}

Our training objective reflects the multi-path nature of the curriculum. For each sample in Stage 2, we optimize over the set of valid reasoning chains $\mathcal{S}_{valid}$. We define the loss as the minimum combined cost across valid paths:
\begin{equation}
\mathcal{L}_{sample} = \min_{s \in \mathcal{S}_{valid}} \left[ \mathcal{L}_{CE}(s) + \gamma \mathcal{L}_{vis}(s) + \eta \mathcal{L}_{p}(s) \right]
\end{equation}
This minimum formulation allows the model to "choose" a path that minimizes the total loss. When multiple paths yield similar cross-entropy loss ($\mathcal{L}_{CE}$) and visual alignment loss ($\mathcal{L}_{vis}$), the sparsity term ($\eta \mathcal{L}_{p}$) naturally biases the selection toward the most concise effective reasoning chain. 

To ensure the observation tokens accurately encode expert knowledge, we employ a composite visual alignment loss:
\begin{equation}
\begin{aligned}[t]
\mathcal{L}_{vis} = \ & \lambda_{seg}\mathcal{L}_{seg} + \lambda_{depth}\mathcal{L}_{depth} \\
& + \lambda_{edge}\mathcal{L}_{edge} + \lambda_{patch}\mathcal{L}_{patch}
\end{aligned}
\end{equation}
where $\mathcal{L}_{seg}$ utilizes Dice and Focal loss with Hungarian matching, $\mathcal{L}_{depth}$ and $\mathcal{L}_{edge}$ use L1 loss, and $\mathcal{L}_{patch}$ uses MSE loss. We set all weights $\lambda$ and $\gamma$ to 1.0. The sparsity weight $\eta$ is set to 0 during Stage 1 and 0.1 during Stage 2 to activate the policy optimization.

\begin{table*}[t]
\centering
\caption{Performance comparison on vision-centric benchmarks. \textbf{Bold} highlights the best result among open-source models. \underline{Underline} indicates the best baseline result per metric.}
\label{tab:main_results}
\begin{threeparttable}
\small
\renewcommand{\arraystretch}{1.2}
\setlength{\tabcolsep}{5pt}
\begin{tabular}{l | ccccccc | c}
\toprule
\textbf{Model \& Paradigm} & \textbf{CV-Bench} & \textbf{BLINK} & \textbf{RW-QA} & \textbf{MMVP} & \textbf{MMStar-P} & \textbf{HR$_{4K}$} & \textbf{HR$_{8K}$} & \textbf{Avg.} \\
\midrule
\rowcolor[gray]{0.9} \textit{Standard Baselines (Backbone)} & & & & & & & & \\
Qwen2.5-VL-7B & 74.5 & 55.7 & 68.6 & 56.0 & 67.1 & 68.6 & 64.9 & 65.1 \\
\quad + Textual CoT Reasoning & 71.2 & 51.8 & 68.9 & 51.5 & 63.5 & 64.9 & 61.2 & 61.9 \\
\midrule
\rowcolor[gray]{0.9} \textit{Visual Reasoning Methods} & & & & & & & & \\
Visual CoT~\cite{shao2024visualcotadvancingmultimodal} (NIPS'24) & 76.8 & 53.5 & 70.1 & 54.2 & 66.8 & 70.5 & 66.3 & 65.5 \\
ICoT$^\ddagger$~\cite{gao2025interleaved} (CVPR'25) & 76.7 & \underline{57.8} & 70.3 & 58.8 & 68.5 & 71.2 & 67.4 & 67.2 \\
Aurora~\cite{bigverdi2025perception} (CVPR'25) & 77.0 & 57.5 & 70.0 & 58.5 & 68.8 & 71.5 & 67.5 & 67.3 \\
CoVT~\cite{qin2025covt} (Pre-print) & \underline{80.0} & 56.0 & 71.6 & 58.7 & 69.2 & 72.9 & 69.4 & 68.3 \\
MINT-CoT~\cite{chen2025mintcot} (NIPS'25) & 78.3 & 57.3 & \underline{72.5} & \underline{60.1} & \underline{70.2} & \underline{73.8} & \underline{70.2} & \underline{68.9} \\
\midrule
\rowcolor[gray]{0.95} \textbf{ViThinker (Ours)} & \textbf{81.4} & \textbf{59.1} & \textbf{74.2} & \textbf{61.3} & \textbf{71.5} & \textbf{76.2} & \textbf{72.5} & \textbf{70.9} \\
\bottomrule
\end{tabular}
\begin{flushleft}
\scriptsize All methods are adapted to Qwen2.5-VL-7B and trained on ViThinker training data, except ICoT$^\ddagger$ which is training-free.
\end{flushleft}
\end{threeparttable}
\vspace{-3mm}
\end{table*}

\section{Experiments}

\subsection{Experimental Setup}

\subsubsection{Implementation}
We implement ViThinker on Qwen2.5-VL-7B~\cite{bai2025qwen25vltechnicalreport} using LoRA~\cite{hu2021lora} (rank 16, alpha 32) for efficient fine-tuning. The embedding layer, language model head, and projection layers are trained with full parameters, while visual experts (SAM~\cite{kirillov2023segment}, DepthAnything~\cite{yang2024depthv2}, PIDINet~\cite{su2021pixeldifferencenetworksefficient}, DINOv2~\cite{oquab2024dinov2learningrobustvisual}) remain frozen. We use AdamW optimizer with learning rates 5e-5 (LoRA) and 1e-5 (projection layers), batch size 4 per GPU, training for 5K steps (Stage 1) and 3K steps (Stage 2) on 2$\times$H100 GPUs.

\subsubsection{Data and Benchmarks}
Our training data combines vision-centric subsets from LLaVA-OneVision~\cite{li2024llavaonevisioneasyvisualtask}, filtered TallyQA~\cite{acharya2018tallyqa}, and ADE20K-Depth~\cite{bigverdi2024perceptiontokensenhancevisual,zhou2018semanticunderstandingscenesade20k}. Stage 1 uses 55k samples with prepended expert outputs; Stage 2 uses 20k interleaved chains with varying query patterns (20\% full, 60\% partial, 20\% minimal).

We evaluate on six vision-centric benchmarks using VLMEvalKit~\cite{vlmevalkit2025}: CV-Bench~\cite{tong2024cambrian1fullyopenvisioncentric}, BLINK~\cite{fu2024blinkmultimodallargelanguage}, MMVP~\cite{tong2024eyeswideshutexploring}, RealWorldQA~\cite{xai2024grok15vision}, MMStar-P~\cite{chen2024rightwayevaluatinglarge}, and HRBench (HR$_{4K}$, HR$_{8K}$)~\cite{HRBench}. Baselines include: (1) Standard VLM (Qwen2.5-VL-7B), (2) Textual CoT~\cite{wei2023chainofthought}, and (3) Sequential Visual Reasoning~\cite{qin2025covt} reproduced on identical data.


\subsection{Main Results}

Table~\ref{tab:main_results} presents our results on vision-centric benchmarks. \method{} consistently outperforms all baselines, establishing a hierarchy of reasoning capabilities driven by perceptual granularity and active engagement.

\subsubsection{Paradigm Comparison.} 
We compare \method{} against four interleaved visual reasoning approaches with increasing sophistication: (1) ICoT~\cite{gao2025interleaved}, a training-free method using attention-based token selection (67.2\% Avg.); (2) Aurora~\cite{bigverdi2024perceptiontokensenhancevisual}, which generates perception tokens via VQVAE for depth maps and coordinates for bounding boxes (67.3\% Avg.); (3) Sequential CoVT, which statically enumerates all dense visual tokens from frozen experts (68.3\% Avg.); (4) MINT-CoT~\cite{chen2025mintcot}, trained with 3-stage curriculum to learn similarity-based token selection (69.3\% Avg.). \method{} achieves +3.6\% over Aurora, +2.6\% over Sequential CoVT, and +2.0\% over MINT-CoT, with particularly pronounced improvements on fine-grained perception tasks (MMVP: +1.2\% vs. MINT-CoT, BLINK: +1.3\% vs. ICoT) and high-resolution benchmarks (HR$_{8K}$: +2.3\% vs. MINT-CoT).


\subsubsection{Analysis of Reasoning Mechanisms.}
The performance hierarchy reveals fundamental limitations in existing approaches: ICoT's attention weights lack semantic precision to distinguish between depth and segmentation; Aurora accumulates VQVAE reconstruction errors (MSE) and is limited to depth/counting tasks, failing on tasks requiring edge detection or semantic understanding; MINT-CoT's similarity-based retrieval passively matches reasoning states to pre-encoded features, unable to dynamically trigger different expert combinations for novel task requirements. In contrast, \method{}'s task-driven query generation explicitly conditions perception on reasoning needs, enabling adaptive visual understanding across diverse modalities. Notably, all internalized visual reasoning methods substantially outperform Textual CoT (-3.2\% vs. Standard Qwen2.5-VL-7B), confirming that grounding reasoning in explicit visual features is crucial for vision-centric tasks.

\begin{figure*}[h]
\centering
\includegraphics[width=0.95\textwidth, height=6cm]{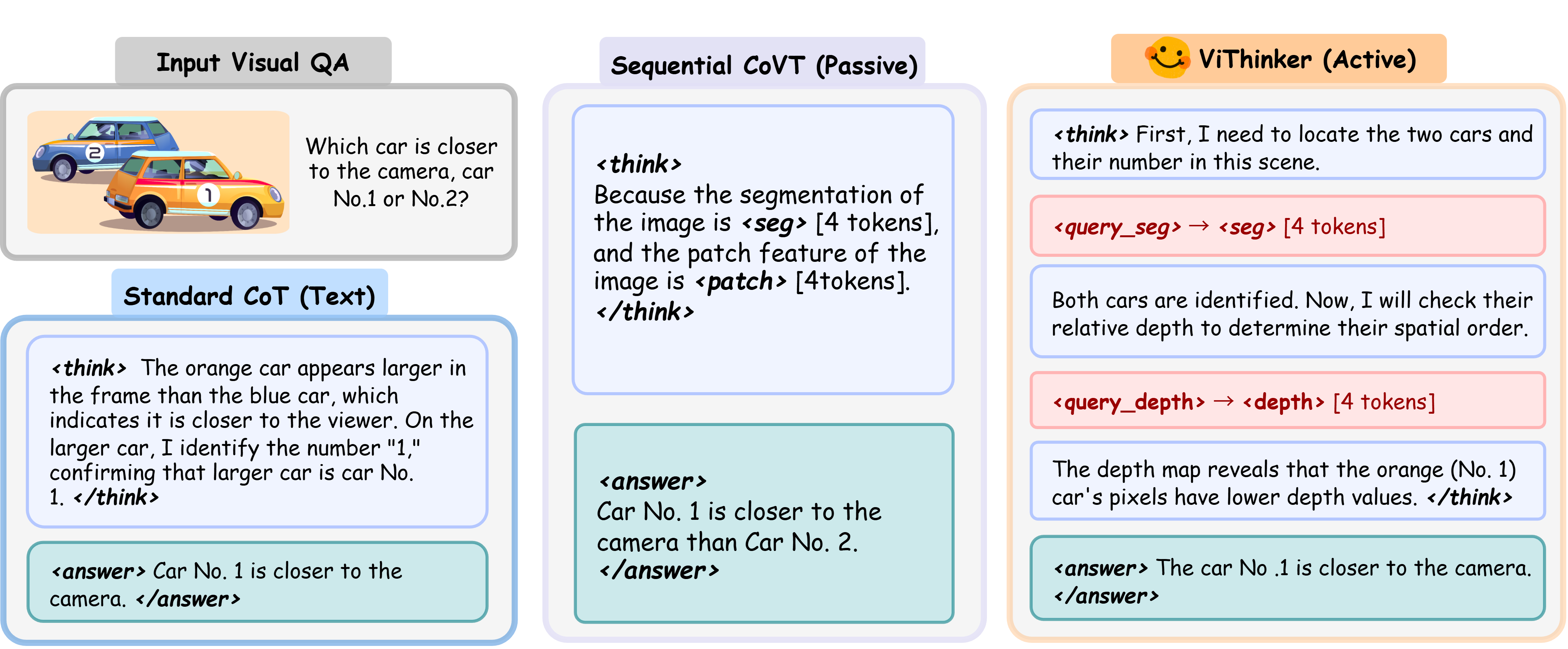}
\caption{\textbf{Qualitative comparison of reasoning paradigms.} Text CoT (left) lacks perceptual grounding. Sequential CoVT (middle) passively generates statistically frequent patterns (seg+patch) learned from training, while ViThinker (right) actively selects task-driven tokens (depth+seg) based on spatial reasoning requirements.}
\label{fig:qualitative}
\end{figure*}

\subsection{Ablation Studies}

\subsubsection{Effect of Training Stages}
We evaluate two-stage curriculum; as shown in Table~\ref{tab:ablation_stages}, both stages are critical. Training with \textbf{Stage 2 only} enables the model to learn interleaved reasoning patterns, but the visual token representations are poorly aligned with perceptual semantics (64.7\% average). \textbf{Stage 1} provides essential foundational grounding by teaching the model to encode perceptual information from frozen experts into visual token representations. The full two-stage curriculum yields a 2.2\% gain over Stage 2 alone, confirming that Stage 1 is a necessary prerequisite for effective reasoning-driven visual token generation (Stage 2).

\begin{table}[h]
\vspace{-2mm}
\centering
\footnotesize
\caption{Ablation on two-stage strategic training curriculum.}
\label{tab:ablation_stages}
\vspace{-4pt}
\begin{tabular}{lcc|c}
\toprule
\textbf{Training Stages} & \textbf{CV-Bench} & \textbf{BLINK} & \textbf{Avg.} \\
\midrule
Stage 2 only & 74.8 & 54.6 & 64.7 \\
Stage 1 + Stage 2 (Full) & \textbf{78.2} & \textbf{57.3} & \textbf{66.9} \\
\midrule
\rowcolor[HTML]{F2F2F2}
$\Delta$ (Stage 1 gain) & +3.4 & +2.7 & +2.2 \\
\bottomrule
\end{tabular}
\vspace{-1mm}
\end{table}

\subsubsection{Qualitative Analysis of Reasoning Paradigms} 
Figure~\ref{fig:qualitative} illustrates the fundamental difference between passive and active reasoning. Methods like Sequential CoVT passively generate statistically frequent token combinations (<seg>, <patch>) learned from training data, regardless of task requirements—exhibiting pattern memorization rather than adaptive selection. In contrast, ViThinker actively adapts its perception to reasoning needs: recognizing this as a spatial comparison task, it generates <query\_seg> for object localization, then conditionally triggers <query\_depth> for spatial relationships. This demonstrates active perceptual reasoning—selecting experts based on what the task requires (depth+seg for spatial comparison) rather than passively enumerating memorized patterns.

\begin{table}[h]
\vspace{-3mm}
\centering
\caption{Reasoning-driven vs. random token generation on CV-Bench.}
\label{tab:efficiency_tradeoff}
\footnotesize
\setlength{\tabcolsep}{5pt}
\begin{tabular}{lcccc}
\toprule
\textbf{Strategy} & \textbf{\#Tokens} & \textbf{2D Tasks} & \textbf{3D Tasks} & \textbf{Overall} \\
\midrule
Full & 16.0 & 83.6 & 79.6 & 81.6 \\
Random & 9.2 & 76.5 & 73.2 & 74.9 \\
\midrule
\rowcolor[HTML]{F2F2F2}
\textbf{\method{}} & \textbf{8.6} & \textbf{83.0} & \textbf{79.8} & \textbf{81.4} \\
\bottomrule
\end{tabular}
\end{table}

\subsubsection{Active vs. Passive Perceptual Selection} 
Table~\ref{tab:efficiency_tradeoff} compares ViThinker's active query generation against passive baselines on CV-Bench: Full Enumeration (passively generating all expert tokens) and Random Pruning (passively masking queries to match token budget). ViThinker significantly outperforms Random Pruning, which degrades performance by inadvertently dropping task-critical experts, demonstrating that passive random selection cannot identify reasoning requirements. Remarkably, ViThinker matches Full Enumeration using 46\% fewer tokens and even outperforms it on 3D tasks (+0.2\%), confirming that passive over-enumeration introduces conflicting perceptual signals. This validates that active, task-driven selection identifies minimal sufficient experts, avoiding both insufficient grounding and noisy over-specification.

\begin{figure}[h]
\centering
\includegraphics[width=0.49\textwidth]{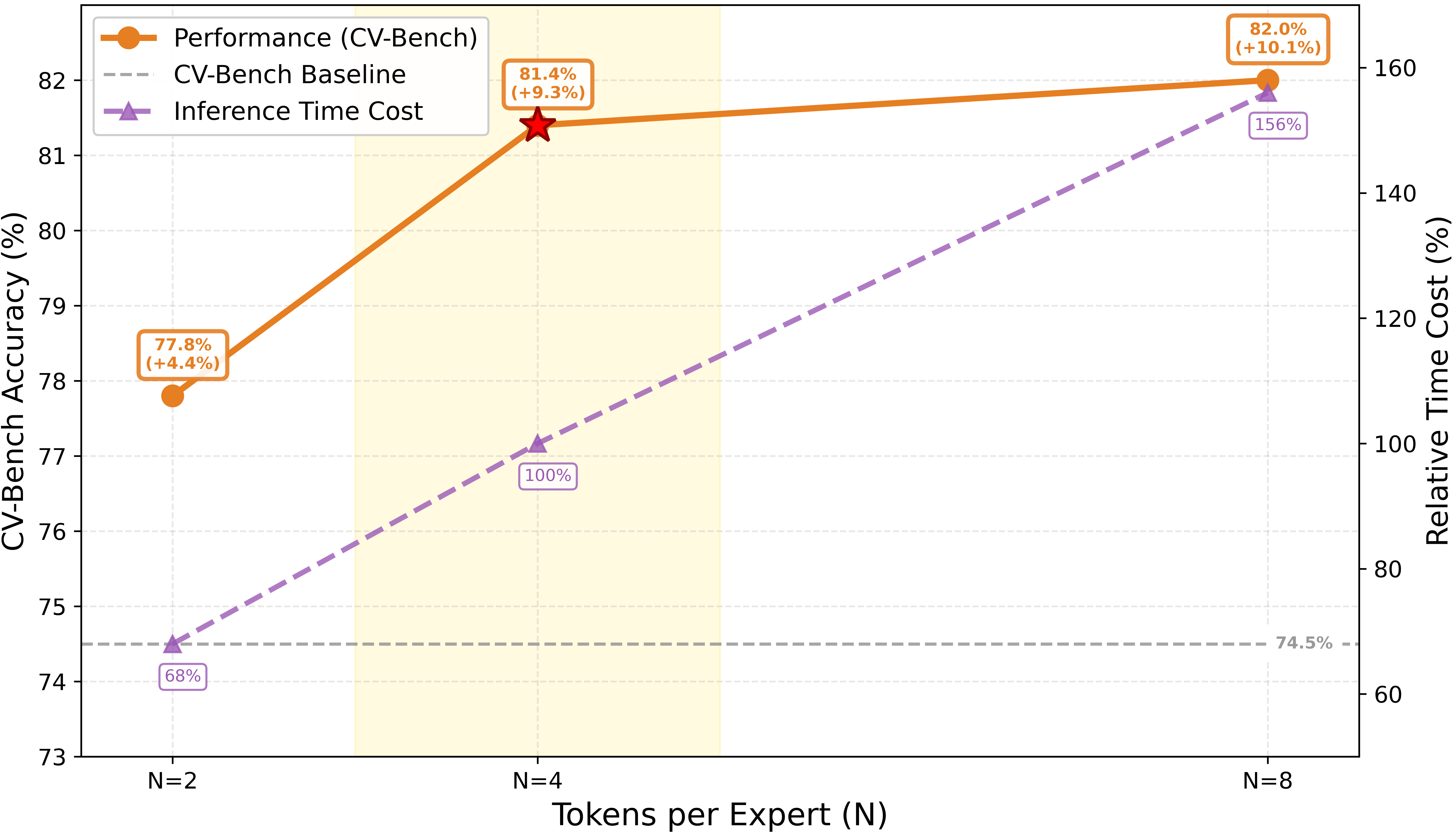}
\caption{Effect of tokens per expert ($N$) on CV-Bench performance (left Y-axis) and inference time cost (right Y-axis).}
\label{fig:tokens_per_expert}
\vspace{-2mm}
\end{figure}

\subsubsection{Tokens per Expert Allocation} 
We ablate the number of observation tokens $N$ per expert during alignment (Eq.~\ref{eq:align}) on CV-Bench to determine the optimal token number configuration. Figure~\ref{fig:tokens_per_expert} shows that $N=2$ achieves 77.8\% due to limited capacity, while $N=4$ reaches 81.4\%. Increasing to $N=8$ yields marginal gains (82.0\%, +0.6\%) at over 50\% higher time cost, validating $N=4$ as the optimal balance. This demonstrates that our configuration (4 tokens) provides sufficient capacity to distill each frozen expert's knowledge, aligning with our principle of minimal sufficient perception.

\begin{figure}[h]
\centering
\includegraphics[width=0.40\textwidth]{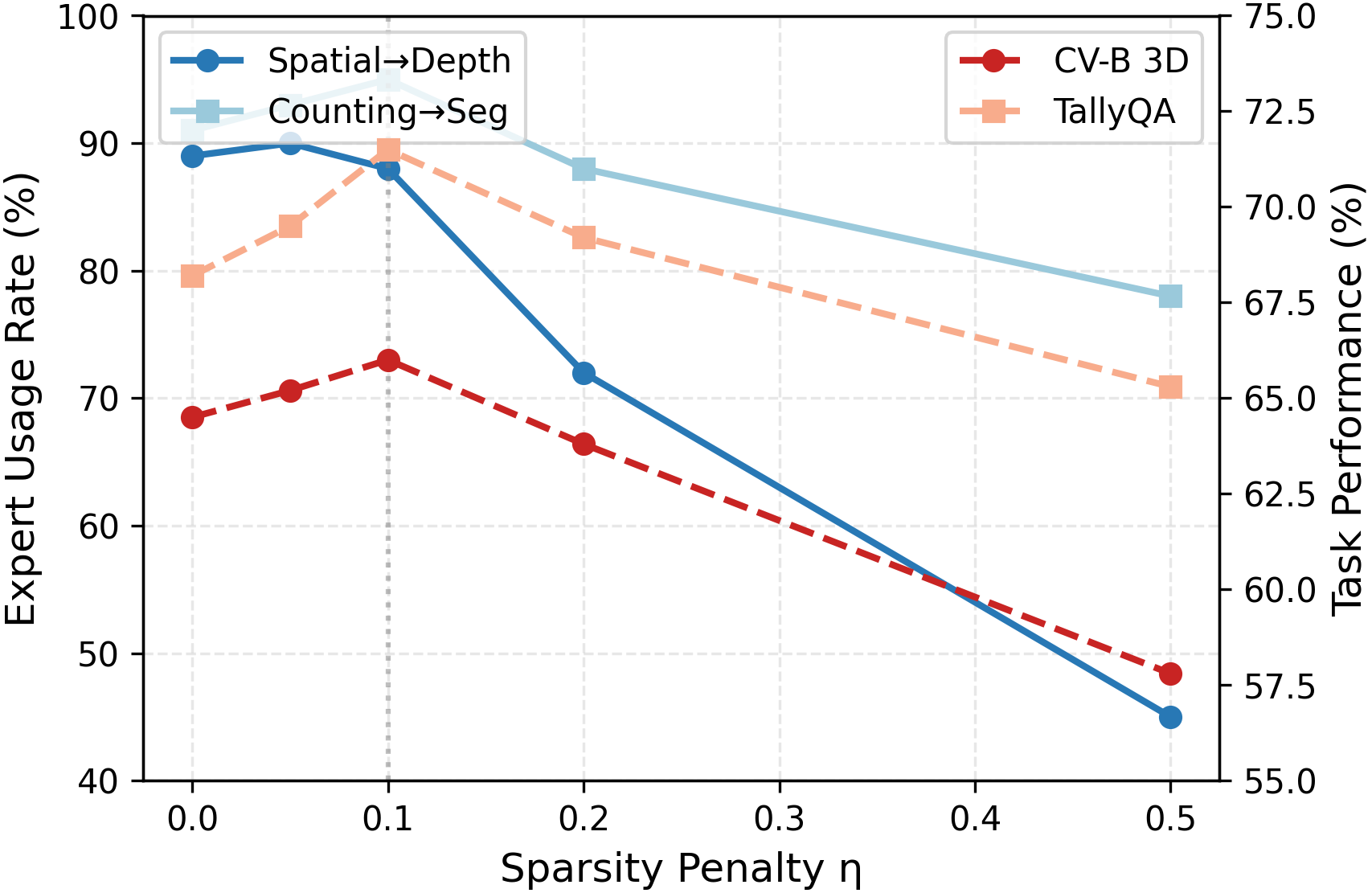}
\caption{Effect of sparsity penalty $\eta$ on task-specific expert selection and performance. Solid lines show expert usage rates (left Y-axis), dashed lines show task performance (right Y-axis).}
\label{fig:eta_ablation}
\vspace{-2mm}
\end{figure}

\subsubsection{Penalty Coefficient Selection} Figure~\ref{fig:eta_ablation} investigates the significance of $\eta$ by task-specific token selection analysis. Without sparsity ($\eta=0$), the model indiscriminately generates all experts, achieving higher usage but suboptimal performance due to lack of selectivity. At $\eta=0.1$, task-appropriate selection peaks: spatial tasks retain depth (88\%) while counting tasks prioritize segmentation (95\%), yielding optimal performance. Excessive penalties ($\eta=0.5$) over-suppress necessary experts, degrading both selection accuracy and task performance. This confirms $\eta=0.1$ encourages task-driven selection.



\section{Conclusion}
We presented ViThinker, a framework enabling active vision-language reasoning through task-driven perceptual querying. By internalizing vision experts during training, ViThinker autonomously synthesizes expert-aligned features at inference without external tool invocation. Our strategic policy learning curriculum teaches the model to trigger perceptual simulations only when reasoning demands them, discovering minimal sufficient perception under sparsity constraints. Empirical results demonstrate consistent improvements in both perceptual grounding and reasoning accuracy over passive baselines. ViThinker establishes a foundation where perception serves not as passive observation, but as dynamic, reasoning-driven action—a key step toward more capable multimodal reasoning systems.



\bibliography{ref}
\bibliographystyle{IEEEbib}

\end{document}